\documentclass[letterpaper, 10 pt, conference]{ieeeconf}  

\IEEEoverridecommandlockouts                              

\usepackage{graphics} 
\usepackage{epsfig} 
\usepackage{mathptmx} 
\usepackage{times} 
\usepackage{amsmath} 
\usepackage{amssymb}  

\usepackage{booktabs}
\usepackage{multirow}
\usepackage{cite}
\usepackage[capitalize]{cleveref}
\usepackage{soul}
\usepackage{color}
\usepackage{threeparttable}

\title{\LARGE \bf
KLTNet: Learning Sparse Feature Tracking for Robust and Accurate Monocular Visual-Inertial Odometry}

\author{Anonymous Author(s)}

\author{Renbiao Jin, Danping Zou\textsuperscript{\dag}, Wenxian Yu %
\thanks{\textsuperscript{\dag}Corresponding Author. All authors are with Shanghai Jiao Tong University, Shanghai 200240, China. (e-mail: \{renbiaojin, dpzou\}@sjtu.edu.cn)}%
}
\begin{document}

\maketitle
\thispagestyle{empty}
\pagestyle{empty}

\begin{abstract}
Many feature-based visual-inertial odometry (VIO) systems rely on sparse feature tracking, whose accuracy and robustness directly affect state estimation. 
Classical KLT trackers rely primarily on local image patches and can become unreliable under rapid motion or in low-texture environments. 
We propose KLTNet, a lightweight learning-based, plug-and-play sparse feature tracker designed to replace classical KLT trackers in KLT-based VIO front ends. 
KLTNet follows a coarse-to-fine, dense-to-sparse architecture that combines low-resolution dense optical flow for robust global motion initialization with triplet-patch refinement for accurate and temporally consistent tracking. A fixed reference patch provides a stable anchor throughout each feature track and helps reduce accumulated tracking drift. 
In addition, KLTNet predicts anisotropic confidence weights supervised through differentiable multi-view triangulation, which can be used as observation weights in compatible VIO estimators. 
Experiments with VINS-Mono and OpenVINS on public benchmarks and a self-collected low-texture dataset demonstrate improved tracking and odometry accuracy over classical KLT, while maintaining real-time performance on an embedded platform.
\end{abstract}

\section{INTRODUCTION}
\begin{figure}[t]
\centering
\includegraphics[width=\columnwidth]{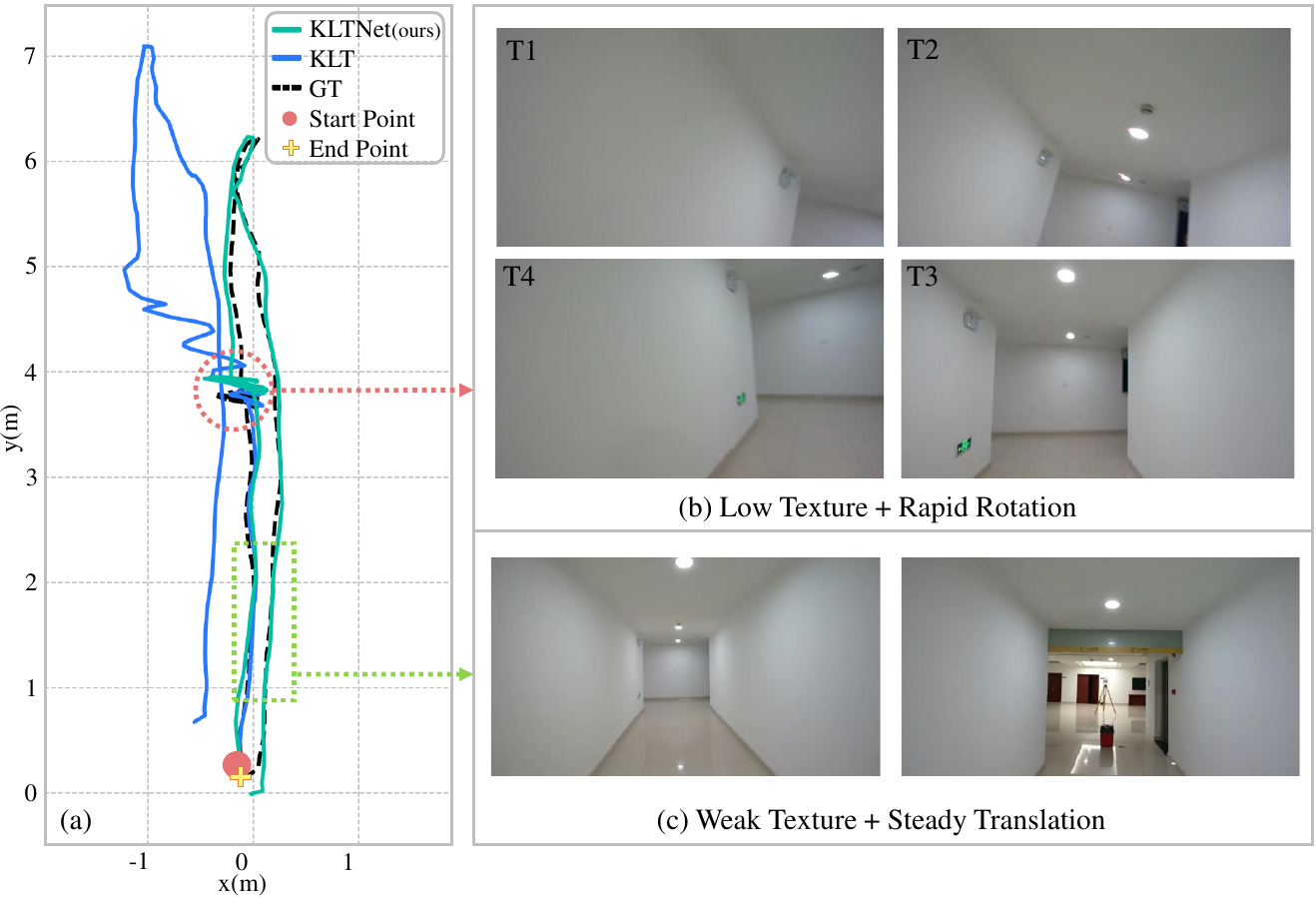}

\caption{
Trajectory comparison in a low-texture environment using VINS-Mono with the classical KLT and the proposed KLTNet tracker. (a) KLTNet closely follows ground truth (GT), while classical KLT drifts under challenging scenarios. (b) During rapid rotation (T1-T4) in low-texture regions, KLT exhibits large drift (red dashed circle), but KLTNet remains stable. (c) In steady translation (green dashed box), both perform similarly, demonstrating that KLTNet performs comparably under moderate motion and remains robust when classical KLT degrades.
}
\vspace{-10pt}
\label{fig:teaser_fig}
\end{figure}

Robust and accurate visual-inertial odometry (VIO) is important for applications such as robotics, autonomous driving, and AR/VR. 
In feature-based VIO systems, sparse feature tracking in the visual front end directly affects the accuracy and robustness of state estimation. The Kanade--Lucas--Tomasi (KLT)\cite{klt91} tracker is widely used because of its efficiency and high frame-to-frame localization accuracy.
However, KLT estimates feature motion mainly from local image patches. Without sufficient image context, tracking can become unreliable under rapid motion or in low-texture environments, such as the challenging corridor example shown in \cref{fig:teaser_fig}.

Recent dense optical flow methods, such as RAFT\cite{raft} and SEA-RAFT\cite{searaft}, achieve robust correspondence estimation by exploiting image-wide context. 
Inspired by these developments, we construct a simple baseline, termed RAFT-Sparse, which propagates sparse feature positions by sampling dense RAFT flow between consecutive frames. However, our experiments show that directly replacing KLT with RAFT-Sparse does not consistently improve odometry accuracy. 
A key limitation is that independently estimated pairwise flow is recursively chained along each feature track, allowing small tracking errors to accumulate over time. These observations suggest that global correspondence robustness alone is insufficient for VIO, where accurate and temporally consistent sparse feature tracks are also important.

Multi-frame learned systems such as DROID-SLAM\cite{droidslam}, DPVO\cite{dpvo} and DVI-SLAM\cite{dvislam} reduce temporal drift by jointly considering visual correspondence and state estimation across multiple frames. However, these methods are designed as complete odometry or SLAM systems in which the learned visual front end is closely coupled with the back end. They are therefore difficult to use as direct replacements for the sparse tracker in existing KLT-based VIO systems and generally introduce additional computational and memory costs. In this work, we instead focus on a standalone sparse tracker that preserves the track-based interface of conventional VIO front ends while improving tracking robustness, accuracy, and temporal consistency.

To this end, we propose KLTNet, a lightweight learning-based, plug-and-play sparse feature tracker designed to replace the classical KLT tracker in KLT-based VIO systems. KLTNet follows a coarse-to-fine, dense-to-sparse architecture. It first estimates low-resolution dense optical flow to capture global motion cues and provide robust initialization for sparse feature tracks. The coarse feature positions are then refined using triplet image patches from the reference, previous, and current frames to improve localization accuracy. Fixing the reference patch at track initialization provides a stable reference throughout the lifetime of a track, helping suppress drift accumulation from recursive frame-to-frame tracking.

Beyond accurate feature tracking, KLTNet also predicts anisotropic confidence weights trained under multi-view geometric constraints through differentiable triangulation. The learned weights provide relative directional weighting for feature observations and can be used for adaptive observation weighting in compatible VIO estimators.

We validate KLTNet by integrating it into VINS-Mono and OpenVINS and evaluating it on multiple public datasets and a self-collected low-texture dataset. 
Experimental results show improved feature tracking accuracy and robustness, as well as lower odometry errors across both VIO systems. KLTNet also achieves real-time tracking on an NVIDIA Jetson AGX Orin.
Our main contributions are summarized as follows:
\begin{itemize}

\item We propose KLTNet, a lightweight learning-based, plug-and-play sparse feature tracker with a coarse-to-fine, dense-to-sparse architecture. It combines global dense-flow initialization with triplet-patch refinement to improve tracking robustness, accuracy, and temporal consistency.

\item We introduce differentiable multi-view triangulation for learning anisotropic observation weights, enabling adaptive feature weighting in compatible VIO estimators.

\item We integrate KLTNet into VINS-Mono and OpenVINS and demonstrate improved odometry accuracy over the classical KLT tracker. With VINS-Mono, KLTNet reduces the average absolute trajectory error by 34\% on EuRoC and 49\% on TUM-VI, while maintaining real-time performance on a resource-constrained platform.
\end{itemize}

\section{Related Work}
\begin{figure*}[t]
\centering
\includegraphics[width=1\textwidth]{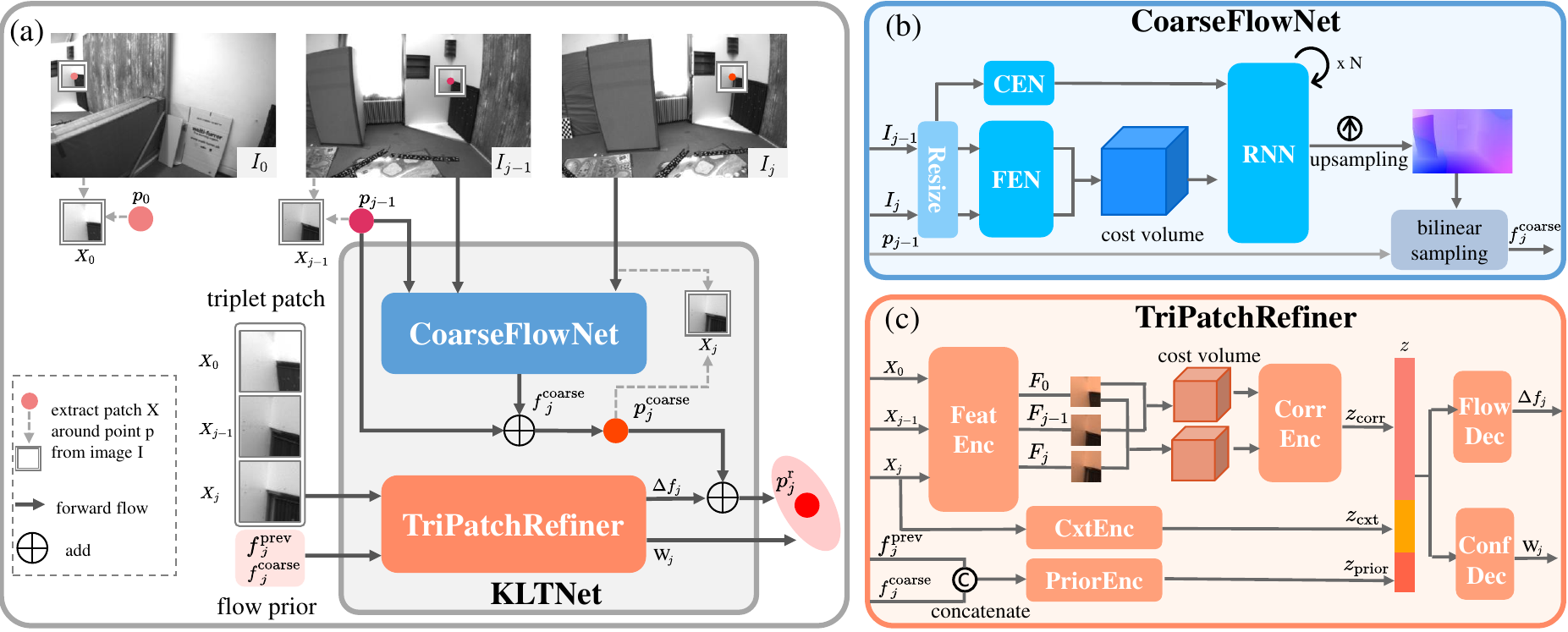}

\caption{
(a) Overview of the KLTNet tracker for sparse feature tracking, illustrated with a single point for clarity. Given a feature $p_0$ with its reference patch $X_0$, and its tracked position $p_{j-1}$ with corresponding patch $X_{j-1}$, \textbf{CoarseFlowNet} first estimates a coarse flow $f_{\text{coarse}}$ between previous frame $I_{j-1}$ and the current frame $I_j$
to predict an initial position $p_j^{\text{coarse}}$. The \textbf{TriPatchRefiner} then processes the triplet image patch$(X_0,X_{j-1},X_j)$, where $X_j$ is extracted around $p_j^{\text{coarse}}$, along with flow priors to regress a fine offset $\Delta f_j$ and a confidence weight matrix $W_j$. The final tracked feature position is obtained by adding the refined offset to the initial prediction.
(b) \textbf{CoarseFlowNet} architecture. It downscales input images, extracts features using FEN and CEN, constructs a cost volume, and iteratively refines flow using an RNN, followed by upsampling and bilinear sampling to get the coarse flow at the tracked point.
(c) \textbf{TriPatchRefiner} architecture. It utilizes FeatEnc to encode triplet image patches. CorrEnc processes cost volumes constructed between patch features. CxtEnc encodes context from patch $X_j$, and PriorEnc encodes flow priors. The concatenated features are fed to FlowDec and ConfDec to predict offset and confidence weights, respectively.
}

\vspace{-15pt}
\label{fig:framework}
\end{figure*} 
Feature-based VIO systems commonly establish visual correspondences through descriptor matching or direct sparse tracking. Descriptor-based systems such as ORB-SLAM3 \cite{orbslam3} detect and match local features across images. In contrast, lightweight VIO systems such as OpenVINS \cite{openvins}, MSCKF \cite{msckf}, and VINS-Mono \cite{vinsmono} commonly use KLT \cite{klt91} to propagate sparse feature points between consecutive frames. KLT is efficient and accurate for frame-to-frame tracking, making it well suited to lightweight VIO front ends.

Several methods improve sparse feature tracking while retaining this formulation. Bouguet \cite{plk} introduces a coarse-to-fine pyramid structure to improve the robustness of KLT under larger motion. LET-VINS \cite{letvins} applies KLT on learned feature maps to relax the brightness constancy assumption. SFVO \cite{sfvo} extracts local patches from dense feature maps and uses CNNs to predict sparse point motion. These methods improve sparse tracking through pyramid processing or learned feature representations. In comparison, KLTNet additionally uses dense optical flow to provide global motion cues before sparse refinement.

Dense optical flow exploits global image context and has achieved substantial improvements with learning-based methods. RAFT \cite{raft} constructs all-pairs correlations and recurrent refinement to estimate dense flow, while SEA-RAFT \cite{searaft} further improves flow accuracy and efficiency. These methods estimate dense correspondence between image pairs and provide strong global motion cues. In KLTNet, dense flow is used only to initialize sparse feature positions, while the final positions are obtained through triplet-patch refinement with a fixed reference patch to improve tracking accuracy and temporal consistency.

Multi-frame visual estimation provides another way to improve temporal consistency. DROID-SLAM \cite{droidslam} jointly optimizes dense flow and bundle adjustment, while DPVO \cite{dpvo} uses sparse deep feature patches to reduce the scale of feature correspondence and bundle adjustment. DVI-SLAM \cite{dvislam} combines learned visual measurements with visual-inertial estimation. These methods improve temporal consistency through joint front-end and back-end estimation, but their visual front ends are closely coupled with the corresponding estimation systems. In contrast, KLTNet is designed as a standalone sparse tracker that can replace the classical KLT tracker in existing KLT-based VIO systems.

Learned observation weighting has also been explored in visual estimation. Uncertainty-aware visual-inertial estimation is studied in \cite{jung2022photometric}, while \cite{dnls} learns correspondence uncertainty through differentiable nonlinear least squares for relative pose estimation. MAC-VO\cite{macvo} predicts metrics-aware covariance for stereo visual odometry. These methods model uncertainty or covariance under their respective estimation formulations. KLTNet instead predicts anisotropic confidence weights for sparse feature observations and supervises them through differentiable multi-view triangulation. The predicted weights are interpreted as anisotropic observation weights rather than metrically calibrated uncertainty, and can be used for adaptive observation weighting in compatible VIO estimators.

\section{Method}
In this section, we introduce KLTNet, a lightweight learning-based, plug-and-play sparse feature tracker designed to replace the classical KLT tracker in KLT-based VIO systems. It consists of a \textbf{CoarseFlowNet} for global robustness and a \textbf{TriPatchRefiner} that improves tracking accuracy. 
Additionally, we introduce a differentiable triangulation strategy to learn confidence weights, which can be used for adaptive observation weighting in compatible VIO estimators.

\subsection{Coarse Dense Flow Initialization}
The classical KLT tracker relies on local image patches, making it susceptible to failure in low-texture environments or rapid motion. In contrast, dense optical flow leverages global image context, providing greater robustness at the expense of higher computational cost. Specifically, the accuracy of flow improves with image resolution, yet the computational cost of constructing the cost volume increases quadratically. Since our method follows a coarse-to-fine architecture, highly accurate flow estimation is not required at the initial stage, allowing us to reduce image resolution and model complexity for efficiency. To this end, we adapt SEA-RAFT into a lightweight network named CoarseFlowNet, specifically designed for low-resolution inputs. We then extract sparse flow from the resulting coarse dense flow as an initial input for subsequent refinement.

Specifically, as shown in Fig. \ref{fig:framework}(b), given two consecutive input images $(I_{j-1}, I_j)$, we first downscale them to 1/4 of their original resolution to reduce computational overhead. A Feature Extraction Network (FEN) is then used to extract 128-channel feature maps at 1/16 of original image resolution. Based on these features, we construct a two-level 4D cost volume by pooling $I_j$'s feature maps at strides of 1 and 2 and computing all-pairs correlations with $I_{j-1}$'s feature maps. Meanwhile, a Context Encoder Network (CEN) takes image $I_{j-1}$ as input and outputs context feature maps, which are injected into a lightweight depthwise-separable convolutional RNN to guide the decoding of the cost volume. The RNN performs four update iterations and outputs the optical flow at 1/16 resolution, which is then upsampled to 1/4 resolution using a learnable convex upsampling module. To initialize feature tracking, we employ bilinear interpolation to extract the sparse optical flow $f_j^{\text{coarse}}$ at specified feature positions $p_{j-1}$ from the previous frame $I_{j-1}$, providing their initial positions in the current frame $I_j$.

\subsection{Fine Sparse Triplet-Patch Refinement}
While dense optical flow provides robust frame-to-frame initialization, recursively propagating pairwise flow estimates can accumulate errors over long feature tracks. Inspired by KLT's local patch alignment, we introduce a triplet-patch refiner that additionally incorporates a fixed reference patch to provide a long-term anchor against drift.

Specifically, as shown in Fig. \ref{fig:framework}(a), when a feature $p_0$ is first detected in frame $I_0$,  an $r\times r$ image patch $X_0$ centered at $p_0$ is extracted and stored as a reference patch. As tracking progresses, in frame $I_{j-1}$, the tracked feature $p_{j-1}$ is associated with a previous patch $X_{j-1}$, extracted in the same manner. For the current frame $I_j$, the initial estimate of the tracked feature $p_j^{\text{coarse}}$ is predicted as $p_{j-1} + f_j^{\text{coarse}}$, where $f_j^{\text{coarse}}$ is the sparse optical flow from CoarseFlowNet, and a current patch $X_j$ is then extracted around $p_j^{\text{coarse}}$. The triplet $(X_0, X_{j-1}, X_j)$ is then encoded by a shared feature encoder (FeatEnc)  into feature maps $(F_0, F_{j-1}, F_j)$. Two 4D cost volumes are computed between the pairs $(F_0,F_{j-1})$ and $(F_0,F_j)$ respectively, which are compressed into a correlation vector $z_{\text{corr}} \in \mathbb{R}^{64}$ via a correlation encoder (CorrEnc). Meanwhile, local appearance and context from $X_j$ are encoded as a context vector $z_{\text{cxt}} \in \mathbb{R}^{32}$ through a context encoder (CxtEnc), while the flow priors $f_j^{\text{coarse}}$ and $f_j^{\text{prev}}$, where $f_j^{\text{prev}} \triangleq p_{j-1} - p_{j-2}$ serves as a historical motion prior, are embedded as $z_{\text{prior}} \in \mathbb{R}^{32}$. These vectors are concatenated and fed into two parallel MLP-based decoders: FlowDec predicts the offset $\Delta f_j$ for the current patch $X_j$, while ConfDec predicts a  positive-definite anisotropic weight matrix $W_j$ associated with the current feature observation. The final refined feature position is $p_j=p_j^{\text{coarse}}+\Delta f_j$.

Since feature-based VIO typically assumes that the tracked features correspond to static scene points, we extract $X_0$ at track initialization and keep it fixed throughout the track's lifetime, rather than continuously updating it as in object tracking. The fixed $X_0$ provides a stable anchor that helps constrain accumulated tracking errors beyond frame-to-frame matching alone. Furthermore, since each feature is tracked independently with its own reference patch, the tracker is compatible with the track-based feature management used by conventional VIO systems.

\subsection{Learning Confidence Weights via Differentiable Triangulation}

To align with the track-based feature management commonly used in conventional VIO, KLTNet predicts per-observation confidence weights and supervises them through differentiable multi-view triangulation within each feature track. Observations of the same feature are coupled through a shared 3D point, while the network predicts a positive-definite $2\times2$ matrix $W$ to modulate their relative contribution and directional weighting.

During training, the tracker is recursively applied over $N$ consecutive frames. For $j=1,\ldots,N-1$, the tracked feature forms the triplet $(X_0,X_{j-1},X_j)$. From each triplet, the TriPatchRefiner parameterized by $\theta$ predicts an offset $\Delta f_{j|\theta}$ and a confidence weight matrix $W_{j|\theta}$, yielding the refined position $p_{j|\theta}=[u_{j|\theta},v_{j|\theta}]^\top$. 
For geometric supervision, we convert the refined positions to normalized image coordinates $x_{j|\theta}$ using the camera intrinsics.
Given camera poses $(R_j,t_j)$, we define the normalized projection matrices $M_j=[R_j|t_j]\in\mathbb{R}^{3\times4}$. We estimate the corresponding 3D point $P_{\theta}$ in two stages.

(1) DLT initialization. Each predicted $p_{j|\theta}$ defines a viewing ray in 3D space, and rays of the same feature track should intersect at a common point. This multi-view constraint is equivalent to the collinearity condition 
$[x_{j|\theta}^\top,1]^\top \times (M_j [P_{\theta}^\top,1]^\top)=0$.
Following the Direct Linear Transform (DLT), expanding the cross product yields two linearly independent constraints. Stacking these equations for all views yields a homogeneous system, which is solved via SVD to obtain an initial estimate $P_{\theta}^{(0)}$.

(2) Confidence-weighted nonlinear refinement. We further refine the 3D point using the weighted reprojection objective:
\begin{equation}
E({P_{\theta}}) = \sum_{j=1}^{N-1}
\left( x_{j|\theta} - \Pi_j(P_{\theta}) \right)^\top
W_{j|\theta}
\left( x_{j|\theta} - \Pi_j(P_{\theta}) \right)
\label{eq:p0_star},
\end{equation}
where $\Pi_j(\cdot)$ denotes perspective projection onto the normalized image plane. Starting from the DLT initialization, we perform two damped Gauss--Newton iterations on \cref{eq:p0_star} and denote the resulting refined point as $P_{\theta}^*$.
The iterations use the normal matrix $\sum_{j}J_j^\top W_{j|\theta}J_j + \lambda I$, where $J_j$ is the reprojection Jacobian and $\lambda=0.1$ is fixed for all training samples. With the damping factor $\lambda$ fixed, globally scaling the predicted weights changes the effective damping of the finite-step solver, making the global scale of the weights identifiable during training.
To obtain a positive-definite confidence weight matrix $W_{j|\theta}$, the raw network outputs $(\tilde{s}, \tilde{\alpha}, \tilde{\beta})$ are mapped to $s = (1+|\tilde{s}|)^{\text{sign}(\tilde{s})}$, $\alpha = \tilde{\alpha}$, and $\beta = (1+e^{-\tilde{\beta}})^{-1}$ as in~\cite{dnls}. We parameterize the weight matrix as:
\begin{equation}
W(s,\alpha,\beta) = s\, R_\alpha \,\text{diag}(\beta,\,1-\beta)\, R_\alpha^\top,
\end{equation}
where $R_{\alpha} \in SO(2)$ is the rotation matrix. 
The predicted confidence weights are learned through differentiable multi-view triangulation and are interpreted as anisotropic observation weights rather than metrically calibrated uncertainty. The predicted weights can therefore downweight observations with unreliable local appearance, such as those affected by viewpoint or illumination changes. 
For downstream VIO, we normalize the predicted scale $s$ by the median scale computed over the training set and discard observations whose normalized scale falls below a fixed empirical threshold or whose refined positions lie outside the image bounds.

The refined 3D point $P_{\theta}^{*}$ is then used to compute the triangulation loss:
\begin{equation}
\mathcal{L}_{\text{tri}}(\theta) = \frac{1}{N-1}\sum_{j=1}^{N-1}\|x_{j|\theta} - \Pi_j(P_{\theta}^{*})\|_1 .
\end{equation}

Because both the predicted feature positions and confidence weights determine $P^{*}_{\theta}$, backpropagating through the differentiable triangulation jointly supervises both outputs. The shared 3D point couples observations within each track and encourages the network to learn observation weighting that is consistent with multi-view geometry. 

\begin{figure*}[t]
\centering

\includegraphics[width=\textwidth]{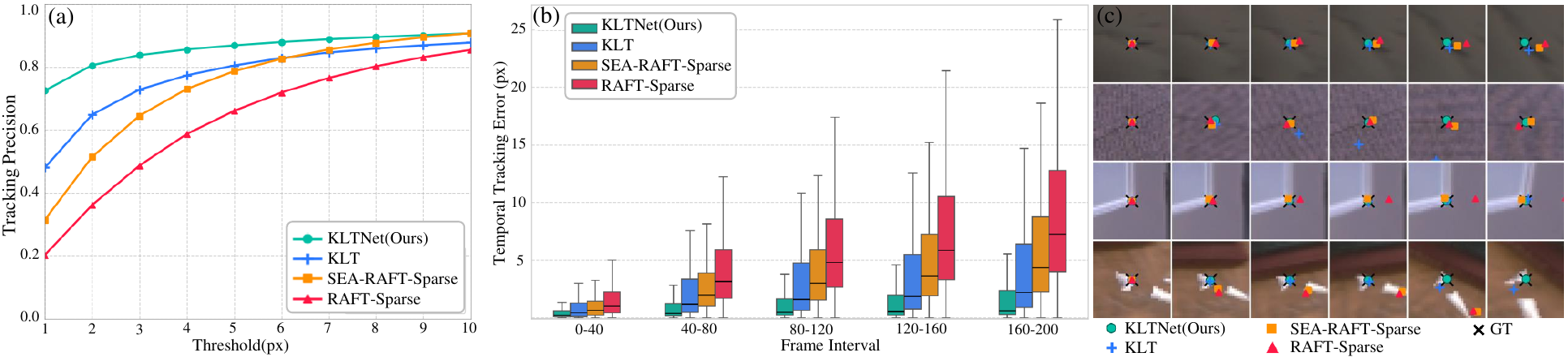}
\caption{
Feature tracking evaluation on the Replica Dataset. (a) Tracking Precision: the percentage of all evaluated feature observations with tracking error below a specified pixel threshold. KLTNet consistently achieves the highest precision, especially at low error thresholds (1--3 pixels). (b) Temporal Tracking Error: distributions of tracking errors across five equal frame intervals of each 200-frame sub-sequence. Box plots show the median and quartiles of the tracking errors. KLTNet exhibits the slowest error growth over time. (c) Qualitative comparison of tracked feature positions across methods.
}

\vspace{-10pt}
\label{fig:track_acc}
\end{figure*}

\subsection{Training and Implementation Details}
\label{subsec:details}
We train KLTNet on the TartanAir\cite{tartanair} dataset in two stages. First, CoarseFlowNet is trained at $1/4$ image resolution following SEA-RAFT\cite{searaft} with the Mixture-of-Laplace loss. Second, CoarseFlowNet is frozen and TriPatchRefiner is trained on 24-frame sequences. Features are detected in the first frame using OpenCV's Good Features to Track (GFTT), and ground-truth feature tracks $p^{\text{gt}}$ are generated using depth and camera poses. Tracks with insufficient accumulated parallax are excluded, resulting in approximately 1.3 million valid tracks.

During TriPatchRefiner training, the refined feature predictions are sequentially propagated throughout the sequence. For each $j=2,\ldots,N-1$, the previous prediction $p_{j-1}$ is detached and used to obtain the coarse position $p_j^{\text{coarse}}$. A fixed $31\times31$ reference patch $X_0$ is extracted at track initialization, while $X_{j-1}$ and $X_j$ are extracted around $p_{j-1}$ and $p_j^{\text{coarse}}$, respectively. TriPatchRefiner then predicts the refinement offset $\Delta f_{j|\theta}$ and confidence weight $W_{j|\theta}$. Sequential propagation exposes the refiner to accumulated tracking errors in the forward pass, while detaching $p_{j-1}$ at each step prevents backpropagation through the full temporal chain. The refinement offset is supervised by $\mathcal{L}_{\text{offset}}$ and the overall objective is:
\begin{equation}
\mathcal{L}(\theta) = 
\underbrace{\frac{1}{N-1}\sum_{j=1}^{N-1}\||(p_j^{\text{gt}} - p_j^{\text{coarse}}) - \Delta f_{j|\theta}||_1}_{\mathcal{L}_{\text{offset}}(\theta)} + \lambda_{\text{tri}} \mathcal{L}_{\text{tri}}(\theta),
\end{equation}
where $\lambda_{\text{tri}}=320$.

\section{Experiments}
In this section, we first evaluate the feature tracking accuracy over consecutive multi-frame sequences. Subsequently, we select VINS-Mono and OpenVINS as two representative monocular VIO systems. We integrate KLTNet into both systems by replacing their native KLT trackers. We validate the effectiveness of our method on multiple public datasets and a self-collected low-texture dataset. 

All methods were evaluated on a system equipped with an Intel i9-10980X CPU, an NVIDIA A6000 GPU, and Ubuntu 20.04. We also evaluated our method on the embedded platform (NVIDIA Jetson AGX Orin) to assess its resource consumption and runtime efficiency.

\subsection{Evaluation of feature tracking}
To evaluate feature tracking accuracy and long-term stability, we conduct experiments on the synthetic dataset Replica\cite{replica}, which comprises 8 indoor scenes with RGB-D sequences of 2000 frames each. For our evaluation, each sequence is divided into 200-frame sub-sequences. Features are initially detected in the first frame of each sub-sequence using the GFTT detector. Ground-truth 2D feature tracks are generated by reprojecting the corresponding 3D points (back-projected using ground-truth depth and camera poses) into each subsequent frame. Ground-truth tracks that become occluded or leave the image boundary are filtered out, resulting in an evaluation set of 5,273 tracks.

We compare KLTNet with three baseline methods: (1) classical KLT tracker as implemented in OpenCV; (2) RAFT-Sparse, which samples sparse flow vectors via bilinear interpolation from the dense flow estimated using the official RAFT checkpoint trained on FlyingChairs and FlyingThings3D; and (3) SEA-RAFT-Sparse, which adopts the same sparse sampling strategy using the  official SEA-RAFT~\cite{searaft} checkpoint trained on TartanAir, FlyingChairs, and FlyingThings3D. RAFT and SEA-RAFT operate at full resolution, whereas KLTNet employs 1/4 resolution for CoarseFlowNet initialization and full resolution for TriPatchRefiner.

At each frame, the tracking error is computed as the Euclidean distance between the feature position propagated by each tracker and its ground-truth position. 
We evaluate performance using two metrics: 
(i) Tracking Precision, the percentage of all evaluated feature observations with error below a given pixel threshold; and (ii) Temporal Tracking Error, obtained by dividing each 200-frame sequence into five equal temporal intervals and aggregating the tracking-error distribution within each interval.

\begin{table*}[t]
\centering
\caption{
Absolute Trajectory Error (ATE) Comparison on the EuRoC Dataset. Results are in meters. Bold indicates best per block.}
\vspace{-5pt}
\begin{threeparttable}
\resizebox{\textwidth}{!}{
\begin{tabular}{c|c|cccccccccccc}
\toprule
\textbf{Method} & \textbf{Tracker} & \textbf{MH01} & \textbf{MH02} & \textbf{MH03} & \textbf{MH04} & \textbf{MH05}  & \textbf{V101}  & \textbf{V102} & \textbf{V103} & \textbf{V201} & \textbf{V202} & \textbf{V203} & \textbf{Average} \\
\midrule
\multirow{4}{*}{VINS-Mono}
                          & KLT       &0.157 &0.179 &0.195 &0.316 &0.302 &0.089 &0.113 &0.188 &0.086 &0.154 &0.277 &0.187 \\
                          & RAFT-Sparse   &0.218 &0.248 &0.224 &0.369 &0.305 &0.076 &0.089 &0.129 &0.110 &\textbf{0.089} &0.236 &0.190 \\
                          & SEA-RAFT-Sparse&0.221 &0.158 &0.223 &0.278 &0.324 &0.098 &0.070 &0.100 &\textbf{0.084} &0.100 &0.166 &0.166 \\ 
                          & KLTNet (Ours) &\textbf{0.092} &\textbf{0.088} &\textbf{0.162} &\textbf{0.233} &\textbf{0.207} &\textbf{0.075} &\textbf{0.070} &\textbf{0.074} &0.089 &0.110 &\textbf{0.149} &\textbf{0.123} \\
\midrule
\multirow{4}{*}{OpenVINS} 
                          & KLT        &0.099 &0.206 &0.150 &0.188 &0.508 &0.069 &0.058 &0.074 &0.105 &\textbf{0.060} &0.122 &0.149 \\
                          & RAFT-Sparse    &0.239 &0.310 &0.235 &0.404 &0.636 &0.073 &0.077 &0.093 &0.112 &0.084 &0.153 &0.219 \\
                          & SEA-RAFT-Sparse &0.238 &0.209 &0.157 &0.399 &0.636 &0.084 &0.068 &0.055 &0.105 &0.074  &\textbf{0.115} &0.194 \\
                          & KLTNet (Ours) &\textbf{0.092} &\textbf{0.099} &\textbf{0.107} &\textbf{0.143} &\textbf{0.187} &\textbf{0.056} &\textbf{0.052} &\textbf{0.062} &\textbf{0.097} &0.064 &0.125 &\textbf{0.099} \\
\midrule
LET-VINS \cite{letvins}   &Learned Feature + KLT &0.195 &0.113 &0.148 &0.192 &0.352 &0.077 &0.121 &0.115 &0.097 &0.150 &0.296 &0.169 \\
DVI-SLAM \cite{dvislam}   &Dense Flow &0.063 &0.083 &0.101 &0.187 &0.163 &0.074 &0.114 &0.083 &0.091 &0.045 &0.072 &0.098  \\
\bottomrule
\end{tabular}
}
\end{threeparttable}
\vspace{-10pt}
\label{tab:euroc}
\end{table*}

\begin{table}[t]
\centering
\caption{
Absolute Trajectory Error (ATE) Comparison on the TUM-VI Dataset. Results are in meters. Bold indicates best per block.
}
\vspace{-5pt}
\resizebox{\columnwidth}{!}{
\begin{tabular}{c|c|ccccccc}
\toprule
\textbf{Method} &\textbf{Tracker} &\textbf{Room1} &\textbf{Room2} &\textbf{Room3} &\textbf{Room4} &\textbf{Room5} &\textbf{Room6} &\textbf{Average}\\
\midrule
\multirow{4}{*}{VINS-Mono } & KLT       &0.067 &0.069 &0.121 &0.047 &0.203 &0.075 &0.097 \\
                           & RAFT-Sparse    &0.120 &0.088 &0.095 &0.074 &0.069 &0.130 &0.096 \\
                           & SEA-RAFT-Sparse&0.067 &0.058 &0.120 &0.047 &\textbf{0.032} &0.109 &0.072 \\
                           & KLTNet (Ours) &\textbf{0.052} &\textbf{0.047} &\textbf{0.061} &\textbf{0.029} &0.081 &\textbf{0.026} &\textbf{0.049} \\
\midrule
\multirow{4}{*}{OpenVINS } &KLT &0.055 &0.103 &0.081 &0.041 &0.067 &\textbf{0.086} &0.072 \\
                          & RAFT-Sparse  &0.116 &0.179 &0.102 &0.098 &0.123 &0.096 &0.119 \\
                          & SEA-RAFT-Sparse&0.086 &0.120 &0.146 &0.029 &0.074 &0.117 &0.095 \\
                          &KLTNet (Ours) &\textbf{0.047} &\textbf{0.060} &\textbf{0.063} &\textbf{0.017} &\textbf{0.064} &0.088 &\textbf{0.057} \\
\midrule
DM-VIO \cite{dmvio} & Photometric      &0.03 &0.13 &0.09 &0.04 &0.06 &0.02 &0.061 \\
\bottomrule
\end{tabular}
}
\vspace{-10pt}
\label{tab:tumvi}
\end{table}

As shown in~\cref{fig:track_acc}(a), KLTNet achieves consistently higher tracking accuracy across all evaluated error thresholds. Notably at low error thresholds (1--3 pixels), KLTNet demonstrates a substantially higher percentage of accurately tracked feature positions compared to the baseline methods. While SEA-RAFT-Sparse improves upon RAFT-Sparse due to its enhanced underlying dense optical flow, it exhibits lower precision than the classical KLT tracker at low error thresholds. However, SEA-RAFT-Sparse's tracking precision begins to surpass that of KLT as the error tolerance is relaxed, suggesting greater robustness against large tracking errors.

\cref{fig:track_acc}(b) illustrates the temporal distribution of tracking errors. Because feature positions are propagated recursively, tracking errors can accumulate over time. However, KLTNet exhibits a substantially slower rate of error accumulation than the baseline methods, as demonstrated by its median error remaining low and stable throughout the sequence. We attribute this improved stability to the fixed reference patch $X_0$, which helps mitigate accumulated track drift. This is further supported by the reference-patch ablation in \cref{tab:ablation_x0}.

\subsection{Evaluation of monocular VIO}
\label{subsec:vio_exp}
To evaluate the odometry accuracy of KLTNet integrated into KLT-based visual-inertial odometry frameworks, we conduct controlled experiments on two public datasets: EuRoC\cite{euroc} and TUM-VI\cite{tumvi}. We analyze three baseline tracking methods: classical KLT, RAFT-Sparse, and SEA-RAFT-Sparse.
We assess KLTNet by integrating it into two distinct VIO systems, VINS-Mono (optimization-based) and OpenVINS (filter-based), replacing their native KLT trackers with the refined feature positions predicted by KLTNet. 
For VINS-Mono, the predicted confidence weights are used to weight the 2D visual reprojection residuals in the normalized image plane, while retaining the original measurement scaling and replacing only the default fixed isotropic weighting.
For OpenVINS, we retain its native measurement-noise model and use the predicted confidence only for observation rejection. 
Loop closure is disabled for all experiments. We report the Absolute Trajectory Error (ATE), calculated as RMSE and averaged over five runs.

\textbf{Results on EuRoC.} As shown in \cref{tab:euroc}, KLTNet demonstrates noticeable performance improvements on the EuRoC dataset.
In the VINS-Mono framework, it achieves a 34\% reduction in average ATE (from $0.187$ to $0.123$) over classical KLT, particularly in challenging sequences such as MH05 and V103. 
SEA-RAFT-Sparse reduces the average ATE by 11\% relative to classical KLT ($0.166$ vs. $0.187$), showing the advantage of learning-based dense flow on more challenging sequences. However, RAFT-Sparse and SEA-RAFT-Sparse still underperform KLT on some sequences because, as \cref{fig:track_acc}(a) shows, dense flow is more robust but less precise than KLT at low error thresholds. Since VIO is sensitive to tracking precision, these small per-frame errors can accumulate over long sequences and lead to larger pose errors, especially on MH01, one of the longest-duration EuRoC sequences with slow camera motion. KLTNet combines the robustness of dense flow with high-precision refinement, while the learned confidence weights provide additional gains through adaptive observation weighting. Their individual contributions are analyzed in \cref{subsec:ablation_study}.
In the OpenVINS framework, KLTNet outperforms the classical KLT by 33\% (from $0.149$ to $0.099$), demonstrating improvements across different VIO systems.

When compared with the additional learned VIO baselines, the monocular OpenVINS integrated with KLTNet demonstrates a favorable balance of accuracy and efficiency. It achieves accuracy comparable to DVI-SLAM---a dense method employing dense optical flow, IMU factors, and bundle adjustment---while outperforming it in computational efficiency (see Sec. \ref{subsec:resource_consumption}).

\textbf{Results on TUM-VI.} For the more challenging TUM-VI handheld dataset experiments (\cref{tab:tumvi}), KLTNet further extends its performance advantages. Integrated with VINS-Mono, it achieves 49\% lower average ATE than classical KLT, maintaining stable tracking in low-texture environments or rapid motion. Using KLTNet tracker, OpenVINS reduces the average ATE by approximately 20\%, whereas using RAFT-Sparse or SEA-RAFT-Sparse leads to performance degradation.

\begin{table}[t]
\centering
\scriptsize
\caption{
Ablation study of KLTNet components integrated with VINS-Mono on EuRoC. Results are the average ATE in meters over the 11 sequences.
}
\vspace{-5pt}
\resizebox{\columnwidth}{!}{
\begin{tabular}{lccllcc}\toprule
\textbf{No.} &\textbf{CoarseFlowNet} &\textbf{TriPatchRefiner} &\textbf{Observation Weight} &\textbf{Weight Supervision} &\textbf{Max Points} &\textbf{Avg.\ ATE} \\\midrule 
R1 &$\checkmark$ &$\times$ &Fixed Isotropic &- &150 &0.317 \\
R2 &$\checkmark$ &$\checkmark$ &Fixed Isotropic &- &150 &0.153 \\
\midrule 
R3 &$\checkmark$ &$\checkmark$ &Isotropic & Reprojection &150 &0.139 \\
R4 &$\checkmark$ &$\checkmark$ &Anisotropic &Reprojection &150 &0.140 \\
R5 &$\checkmark$ &$\checkmark$ &Isotropic &Triangulation &150 &0.131 \\
R6 &$\checkmark$ &$\checkmark$ &Anisotropic &Triangulation &150 &0.123 \\\midrule 
R7 &$\checkmark$ &$\checkmark$ &Anisotropic &Triangulation &100 &0.144 \\
R8 &$\checkmark$ &$\checkmark$ &Anisotropic &Triangulation &50 &0.158 \\
R9 &$\checkmark$ &$\checkmark$ &Anisotropic &Triangulation &25 &0.206 \\\bottomrule
\end{tabular}
}
\vspace{-10pt}
\label{tab:ablation_study}
\end{table}

\begin{table}[t]
\centering
\caption{Impact of the reference patch $X_0$ on tracking accuracy (Replica) and VIO accuracy (EuRoC).}
\vspace{-5pt}
\label{tab:ablation_x0}
\resizebox{\columnwidth}{!}{
\begin{tabular}{@{}lcccc@{}}
\toprule
\multirow{2}{*}{\textbf{Input Patches}} & \multicolumn{2}{c}{\textbf{Median Tracking Error (px)}} & \multicolumn{2}{c}{\textbf{VIO ATE (m)}} \\
\cmidrule(lr){2-3} \cmidrule(lr){4-5}
&Short interval & Long interval & VINS-Mono & OpenVINS \\
\midrule
two-patch ($X_{j-1}, X_j$)      & 0.83 & 1.22 & 0.135 & 0.134 \\
triplet-patch ($X_0, X_{j-1}, X_j$) & 0.25 & 0.59 & 0.123 & 0.099 \\
\bottomrule
\end{tabular}
}
\vspace{-10pt}
\end{table}

\begin{table*}[t]
\centering
\caption{
Tracker comparison within VINS-Mono on KITTI-360 and the illumination-change subset of UMA-VI.
KITTI-360: $t_{rel}(\%)$ / $r_{rel}(^\circ /100m)$;
UMA-VI: start-to-end errors in meters.
Best results are in bold.
}
\label{tab:kitti_uma}
\vspace{-5pt}
\scriptsize
\setlength{\tabcolsep}{2.2pt}
\renewcommand{\arraystretch}{0.92}

\resizebox{\textwidth}{!}{
\begin{tabular}{c|cccccccc|ccccc}
\toprule
& \multicolumn{8}{c|}{\textbf{KITTI-360}}
& \multicolumn{5}{c}{\textbf{UMA-VI}} \\
\textbf{Tracker}
& \textbf{00} & \textbf{02} & \textbf{03} & \textbf{04}
& \textbf{05} & \textbf{06} & \textbf{09}$^*$ & \textbf{10}
& \textbf{C1} & \textbf{C2} & \textbf{C3} & \textbf{T1} & \textbf{T2} \\
\midrule

KLT & 1.17/0.15 & 1.10/0.22 & 1.33/\textbf{0.10} & 1.02/0.21 & 1.02/0.26 & 0.99/0.17 & 1.06/0.14 & 2.00/0.24 & 4.50 & 1.41 & 0.81 & 0.24 & 0.22 \\

SEA-RAFT-Sparse & 1.99/0.21 & 1.30/0.27 & 1.08/0.15 & 1.86/0.24 & 1.09/0.23 & 1.34/0.21 & 1.60/0.20 & 1.88/0.27 & 7.89 & 1.07 & 0.18 & 0.27 & 0.41 \\

KLTNet (Ours) & \textbf{0.72}/\textbf{0.12} & \textbf{0.65}/\textbf{0.19} & \textbf{0.78}/0.11 & \textbf{0.82}/\textbf{0.18} & \textbf{0.88}/\textbf{0.22} & \textbf{0.81}/\textbf{0.15} & \textbf{0.69}/\textbf{0.13} & \textbf{1.65}/\textbf{0.21} & \textbf{1.93} & \textbf{0.49} & \textbf{0.16} & \textbf{0.18} & \textbf{0.16} \\

\bottomrule
\end{tabular}
}

\begin{tablenotes}
\footnotesize
\item $^*$ Only the first 923 s are used due to an IMU data gap.
C1--C3: conference-csc1--3; T1--T2: third-floor-csc1--2.
\end{tablenotes}

\vspace{-10pt}
\end{table*}

\subsection{Ablation Study}
\label{subsec:ablation_study}
To further analyze the contribution of the major components, we conduct an ablation study on the EuRoC dataset by integrating KLTNet into the VINS-Mono framework.

\textbf{Impact of the Patch Refinement.} 
To evaluate the effect of patch refinement, \cref{tab:ablation_study}-R1 and \cref{tab:ablation_study}-R2 both use VINS-Mono's default fixed weights. \cref{tab:ablation_study}-R1 uses only CoarseFlowNet, which preserves system stability but yields imprecise tracking and a higher ATE. \cref{tab:ablation_study}-R2 further incorporates TriPatchRefiner, forming the full KLTNet feature tracker. Adding TriPatchRefiner substantially reduces ATE, demonstrating its substantial contribution to the overall tracking performance.

\textbf{Impact of Anisotropic Confidence Weights.} We examine how the representation and supervision of confidence weights jointly affect VIO accuracy. An NLL-based baseline (\cref{tab:ablation_study}-R3) models the 2D position as a Gaussian with scalar variance supervised by a reprojection-only negative log-likelihood against ground truth. 
The predicted variance is converted into an isotropic residual weight for bundle adjustment, yielding a marginal improvement over fixed weighting. Extending the same reprojection-based supervision to an anisotropic weighting formulation (\cref{tab:ablation_study}-R4) yields comparable performance, suggesting that the additional directional degrees of freedom do not translate into measurable VIO gains under this training setup. In comparison, using differentiable multi-view triangulation supervision improves the isotropic formulation (\cref{tab:ablation_study}-R5) over its reprojection-based counterpart, while the anisotropic triangulation variant (\cref{tab:ablation_study}-R6) achieves the lowest ATE, further improving over the fixed-weight KLTNet configuration in \cref{tab:ablation_study}-R2.

\textbf{Impact of the Number of Tracked Points.} We vary the maximum number of tracked feature points per frame from 150 down to 25 (\cref{tab:ablation_study}-R7--R9). As expected, fewer points cause a gradual increase in ATE, but even with as few as 25 points the system still runs reliably and achieves reasonable accuracy, benefiting from KLTNet's strong tracking precision and confidence weights.

\textbf{Impact of the Reference Patch $X_0$.} To isolate the contribution of $X_0$, we compare the full TriPatchRefiner with a two-patch variant using only $X_{j-1}, X_j$ (single cost volume, other components unchanged). \cref{tab:ablation_x0} reports the median tracking error on Replica (short: 0--40 frames; long: 160--200 frames) and the average EuRoC ATE under VINS-Mono and OpenVINS. Removing $X_0$ raises the tracking error at both intervals, and more markedly so at long intervals, confirming that $X_0$ primarily exposes accumulated tracking errors that are not observable from frame-to-frame matching alone. The improvement in tracking consistency translates to lower VIO ATE on both VIO systems.

\begin{figure}[t]
\centering
\includegraphics[width=\columnwidth]{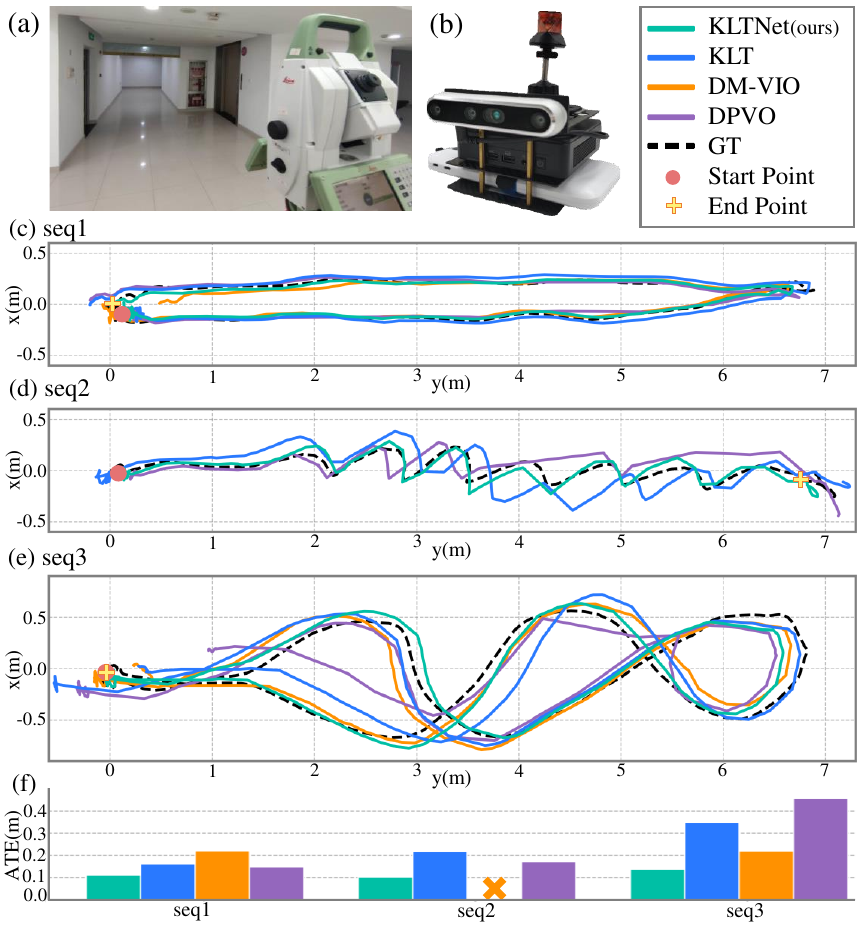}
\vspace{-15pt}
\caption{
VIO performance on a self-collected low-texture dataset. (a) A view of the challenging low-texture environment. (b) The handheld data acquisition platform. (c)-(e) Estimated trajectories for KLTNet-VINS-Mono compared against KLT-VINS-Mono, DM-VIO, and DPVO. (f) Absolute Trajectory Error (ATE) in meters. Note that DM-VIO failed to track in seq2.
}
\vspace{-15pt}
\label{fig:robust_exp}
\end{figure}

\subsection{Robustness Evaluation}
\label{subsec:robustness_exp}
To evaluate robustness beyond indoor benchmarks, we conduct controlled tracker comparisons within VINS-Mono on KITTI-360\cite{kitti360} and the illumination-change subset of UMA-VI\cite{umavi}, using the same VINS-Mono configuration as in \cref{subsec:vio_exp}. As shown in Table V, KLTNet achieves the lowest relative translation error on all eight KITTI-360 sequences and the lowest relative rotation error on seven, reducing the mean trel by 27.8\% over KLT. On UMA-VI, where ground truth is available only at the beginning and end of each sequence, KLTNet obtains the lowest start-to-end error on all five sequences. These results demonstrate robust generalization to long outdoor trajectories and abrupt illumination changes. 
We next focus on the low-texture indoor setting, where classical KLT tracking can degrade because local image structure is limited. We collected a dataset in a low-texture indoor corridor with uniform white walls (Fig. \ref{fig:robust_exp}(a)). Data was acquired using a handheld platform (Fig. \ref{fig:robust_exp}(b)) equipped with an Intel RealSense D455 camera (monocular + IMU). Ground-truth trajectories were obtained using a total station. We integrated our tracker into VINS-Mono (denoted KLTNet-VINS-Mono) and benchmarked it against the original system (KLT-VINS-Mono), DM-VIO, and DPVO.

As shown by the ATE in Fig. \ref{fig:robust_exp}(f), KLTNet-VINS-Mono consistently achieves the lowest error across all sequences. While most methods perform adequately in the simple back-and-forth motion of seq1 (Fig. \ref{fig:robust_exp}(c)), the advantages of KLTNet become evident as motion complexity increases. In seq2 (Fig. \ref{fig:robust_exp}(d)), which involves forward translation with continuous pitch and roll rotations, DM-VIO fails to track, and the baseline KLT-VINS-Mono accumulates substantial drift. Similarly, during the complex serpentine motion with sharp turns in seq3 (Fig. \ref{fig:robust_exp}(e)), both KLT-VINS-Mono and DPVO exhibit large drift. In contrast, KLTNet-VINS-Mono maintains stable and accurate tracking in all cases.

\begin{figure}[t]
\centering
\includegraphics[width=0.9\columnwidth]{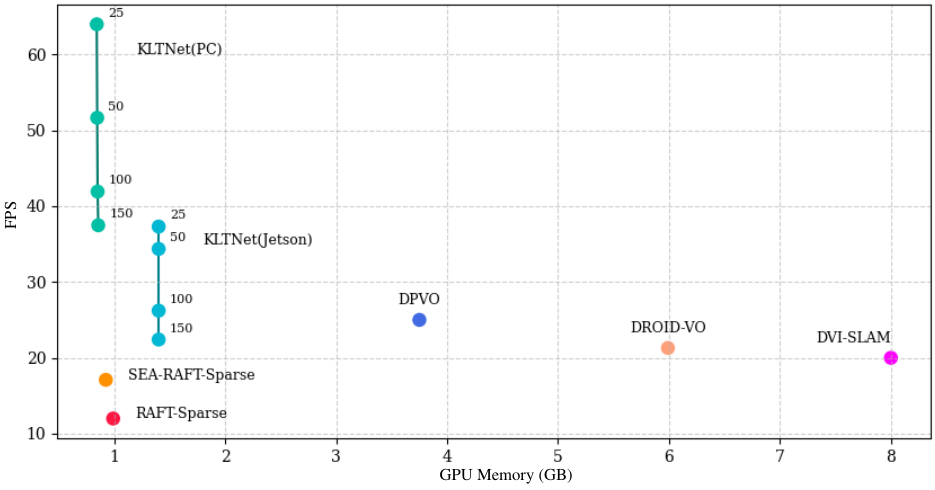}
\vspace{-10pt}
\caption{
Throughput and GPU Memory Comparison. KLTNet, RAFT-Sparse, and SEA-RAFT-Sparse are evaluated within the VINS-Mono visual front end, with throughput measured for the complete front-end pipeline. KLTNet is evaluated on a PC (NVIDIA A6000; LibTorch without TensorRT) and a Jetson AGX Orin (with TensorRT) with maximum track counts of 25, 50, 100, and 150. End-to-end learned odometry systems are shown as system-level reference points.}
\vspace{-20pt}
\label{fig:resource_exp}
\end{figure}

\subsection{Resource Consumption}
\label{subsec:resource_consumption}
Practical VIO deployment, especially on embedded platforms, demands an efficient visual front end. We therefore evaluate the runtime and GPU memory consumption of the VINS-Mono visual front end using KLTNet, RAFT-Sparse, and SEA-RAFT-Sparse as the feature tracker. \cref{fig:resource_exp} reports results for KLTNet on both a PC (NVIDIA A6000) and an embedded Jetson AGX Orin with different maximum track counts. KLTNet runs in C++ with LibTorch on the PC without TensorRT acceleration, while CoarseFlowNet and TriPatchRefiner are accelerated with TensorRT on the Jetson. RAFT-Sparse and SEA-RAFT-Sparse are included as standalone learned tracking baselines. End-to-end learned odometry systems, including DROID-VO (DROID-SLAM in odometry mode), DPVO, and DVI-SLAM, are also shown as system-level reference points rather than direct tracker baselines. KLTNet achieves real-time tracking on the Jetson AGX Orin with low GPU memory consumption, supporting its use in resource-constrained VIO deployments.

\section{CONCLUSIONS}
We present KLTNet, a lightweight, learning-based, plug-and-play sparse feature tracker designed to replace the classical KLT tracker. Its coarse-to-fine architecture combines robust dense flow initialization with precise triplet-patch refinement. A differentiable triangulation strategy learns confidence weights, which can be used for adaptive observation weighting in compatible KLT-based VIO systems. Experiments across multiple datasets and VIO frameworks demonstrate improved accuracy and robustness while maintaining real-time performance on embedded platforms. As a coarse-to-fine tracker, KLTNet relies on the coarse flow to initialize features within the capture range of the local refinement stage, and the current formulation assumes predominantly static scene points. Future work will explore more robust re-initialization and dynamic-scene handling by exploiting the dense motion cues already available from CoarseFlowNet, as well as extending KLTNet to multi-camera systems.


\bibliographystyle{IEEEtran}
\bibliography{IEEEabrv}

\end{document}